\documentclass[a4paper,fleqn]{cas-sc}

\usepackage[authoryear]{natbib}
\def\tsc#1{\csdef{#1}{\textsc{\lowercase{#1}}\xspace}}
\tsc{WGM}
\tsc{QE}
\tsc{EP}
\tsc{PMS}
\tsc{BEC}
\tsc{DE}
\begin{document}
\let\WriteBookmarks\relax
\def\floatpagepagefraction{1}
\def\textpagefraction{.001}

% Short title
\shorttitle{Machine Learning and Computer Vision Techniques in Continuous Beehive Monitoring Applications: A Survey}

% Short author
\shortauthors{S Bilik et~al.}

% Main title of the paper
\title [mode = title]{Machine Learning and Computer Vision Techniques in Continuous Beehive Monitoring Applications: A Survey}                      
% Title footnote mark
% eg: \tnotemark[1]
% \tnotemark[1,2]

% Title footnote 1.
% eg: \tnotetext[1]{Title footnote text}
% \tnotetext[<tnote number>]{<tnote text>} 
% \tnotetext[1]{This document is the results of the research
%    project funded by the National Science Foundation.}

% \tnotetext[2]{The second title footnote which is a longer text matter
%    to fill through the whole text width and overflow into
%    another line in the footnotes area of the first page.}

% First author
%
% Options: Use if required
% eg: \author[1,3]{Author Name}[type=editor,
%       style=chinese,
%       auid=000,
%       bioid=1,
%       prefix=Sir,
%       orcid=0000-0000-0000-0000,
%       facebook=<facebook id>,
%       twitter=<twitter id>,
%       linkedin=<linkedin id>,
%       gplus=<gplus id>]

\author[1,2]{Simon Bilik}[
                        %type=editor,
                        auid=000,bioid=1,
                        %role=Researcher,
                        orcid=0000-0001-8797-7700]

% Corresponding author indication
\cormark[1]

% Email id of the first author
\ead{bilik@vut.cz}

%  Credit authorship
\credit{Conceptualization of this study (lead), Methodology (equal), Writing – Original Draft Preparation (equal)}

\author[1]{Tomas Zemcik}[
                        %type=editor,
                        auid=001,bioid=2,
                        %role=Researcher,
                        orcid=0000-0003-4363-4313]

%  Credit authorship
\credit{Investigation (equal), Methodology, Writing – Original Draft Preparation (equal)}

\author[1]{Lukas Kratochvila}[
                        %type=editor,
                        auid=003,bioid=3,
                        %role=Researcher,
                        orcid=0000-0001-8425-323X]

%  Credit authorship
\credit{Investigation (equal), Writing – Original Draft Preparation (equal)}

\author[1]{Dominik Ricanek}[
                        %type=editor,
                        auid=004,bioid=4,
                        %role=Researcher,
                        orcid=0000-0001-5031-2481]

%  Credit authorship
\credit{Writing – Review \& Editing (equal)}

\author[1]{Miloslav Richter}[
                        %type=editor,
                        auid=006,bioid=6,
                        %role=Researcher,
                        orcid=0000-0002-9791-5957]

%  Credit authorship
\credit{Writing – Review \& Editing (equal)}

\author[3]{Sebastian Zambanini}[
                        %type=editor,
                        auid=007,bioid=7,
                        %role=Researcher,
                        orcid=0000-0002-3459-8122]

%  Credit authorship
\credit{Supervision (equal), Writing – Review \& Editing (equal)}

\author[1]{Karel Horak}[
                        %type=editor,
                        auid=005,bioid=5,
                        %role=Researcher,
                        orcid=0000-0002-2280-3029]

%  Credit authorship
\credit{Supervision (equal), Writing – Review \& Editing (equal)}

% Footnote of the first author
%\fnmark[1]

% URL of the first author
%\ead[url]{www.cvr.cc, cvr@sayahna.org}

% Address/affiliation
\affiliation[1]{organization={Department of Control and Instrumentation, Faculty of Electrical Engineering and Communication, Brno University of Technology},
    addressline={Technická 3058/10}, 
    city={Brno},
    % citysep={}, % Uncomment if no comma needed between city and postcode
    postcode={61600}, 
    % state={},
    country={Czech Republic}}

\affiliation[2]{organization={Computer Vision and Pattern Recognition Laboratory, Department of Computational Engineering, Lappeenranta-Lahti University of Technology LUT},
    addressline={Yliopistonkatu 34}, 
    city={Lappeenranta},
    % citysep={}, % Uncomment if no comma needed between city and postcode
    postcode={53850}, 
    % state={},
    country={Finland}}

\affiliation[3]{organization={Computer Vision Lab, Institute of Visual Computing \& Human-Centered Technology, Faculty of Informatics, TU Wien},
    addressline={Favoritenstr. 9/193-1}, 
    city={Vienna},
    % citysep={}, % Uncomment if no comma needed between city and postcode
    postcode={A-1040}, 
    % state={},
    country={Austria}}

% Corresponding author text
\cortext[cor1]{Corresponding author}
% \cortext[cor2]{Principal corresponding author}

% Footnote text
% \fntext[fn1]{This is the first author footnote. but is common to third
%   author as well.}
% \fntext[fn2]{Another author footnote, this is a very long footnote and
%   it should be a really long footnote. But this footnote is not yet
%   sufficiently long enough to make two lines of footnote text.}

% For a title note without a number/mark
% \nonumnote{This note has no numbers. In this work we demonstrate $a_b$
%   the formation Y\_1 of a new type of polariton on the interface
%   between a cuprous oxide slab and a polystyrene micro-sphere placed
%   on the slab.
%   }

% Here goes the abstract
\begin{abstract}
    Wide use and availability of machine learning and computer vision techniques allows development of relatively complex monitoring systems in many domains. Besides the traditional industrial domain, new applications appears also in biology and agriculture, where they may be used to detect infections, parasites and weeds, but also for automated monitoring and early warning systems. This goes in concordance with the introduction of the easily accessible hardware and development kits such as the Arduino, or RaspberryPi families. In this paper, we survey 50 papers focusing on the methods of automated beehive monitoring using computer vision techniques. Particularly on the pollen and Varroa mite detection together with the bee traffic monitoring. Such systems could also be used for monitoring of honeybee colonies and for the inspection of their health state, which could potentially identify dangerous states before the situation is critical, or to better plan periodic bee colony inspections and therefore save significant costs. Further on, we also include analysis of the research trends in this application field and we outline the possible directions of new development. Our paper is also aimed at veterinary and apidology professionals and experts, who may not be familiar with machine learning to introduce them to its capabilities, hence each family of techniques is prefaced by a brief theoretical introduction and motivation related to its base method. We hope that this paper will inspire other scientists to use machine learning techniques for other applications in beehive monitoring.
\end{abstract}

% Use if graphical abstract is present
% \begin{graphicalabstract}
% \includegraphics{figs/grabs.pdf}
% \end{graphicalabstract}

% Research highlights
%\begin{highlights}
%\item Computer vision techniques could be used for the automated beehive monitoring.
%\item Research in this field starts to focus on the object detection techniques.
%\item Bee traffic monitoring research grows rapidly using the state-of-the-art techniques.
%\end{highlights}

% Keywords
% Each keyword is seperated by \sep
\begin{keywords}
\sep Pollen detection \sep Varroasis detection \sep Bee traffic inspection \sep Bee inspection \sep Machine learning  \sep Computer vision
\end{keywords}

\maketitle

\section{Introduction}\label{Sect:Intro}

The honey bee (\textit{Apis mellifera}) is one of the most important pollinators worldwide and it covers a significant part of the floral visits, some of them exclusively \cite{doi:10.1098/rspb.2017.2140}. Nevertheless, honey bee colonies worldwide face numerous problems connected to parasites such as the Varroa mite (\textit{Varroa destructor}), viral diseases, and colony collapse, which, if left unchecked, might cause problems with the food supply chains and huge economic losses, not to mention other related environmental damage. This problem affects both wild and bred pollinators with the expected losses from 3\% to 8\% of the agricultural production in the case of the total absence of animal pollinators \cite{10.1093/aob/mcp076}.

In recent years, modern machine learning techniques have proven to be very efficient for processing multidimensional data with a huge amount of information and dependencies. These techniques, which are often used in computer vision applications, made a breakthrough in previously hard-to-solve problems, such as classification, recognition, or inspection tasks, and they could be successfully applied to automated bee inspection applications. For example, a comprehensive overview of bee counting methods developed over the last century is presented in \cite{https://doi.org/10.1111/aab.12727}, but despite the high quality of this paper, the machine learning-based methods and other applications are mentioned only briefly. We aim to bridge the existing gap with this paper, because a wider use of these techniques could lead to time and economic savings in bee colony research, or to early recognition of some potentially dangerous situations, possibly allowing for a corrective action before dramatic measures are required. Most of the described techniques could be easily used for recognition of pollen-bearing bees, bee counting, foreign insect detection, or early diagnosis of various infections.

For the above-mentioned reasons, this paper presents the state-of-the-art methods for beehive monitoring based on machine learning, computer vision or their combination in more detail. In Section \ref{Sect:CV-DS} of our paper, we briefly describe the basic concepts of machine learning and deep learning followed by a brief introduction of the conventional computer vision and machine learning based methods, CNN-classifier based methods and object detector based methods followed by an overview of available bee datasets. This part is followed by Section \ref{Sect:Applic} with a survey of existing papers in pollen detection, Varroa mite detection, bee traffic monitoring and general bee inspection application fields. In Section \ref{Sect:Disc}, we discuss trends in automated beehive monitoring and possible future research directions. We believe that this paper will aid the introduction of these techniques into everyday use and the presentation of their potential to researchers from non-technical fields.

\section{Computer vision methods and machine learning based beehive monitoring applications}\label{Sect:CV-DS}

Machine learning is a part of the so-called artificial intelligence, which is a term covering a broad range of methods, models, and algorithms, that build computer models for various applications in fields from computer science (computer vision), classification and recognition tasks, physics (modelling physical processes), pharmacology, and biology (discovering new drugs and molecules). The term artificial intelligence has many definitions, and it is often misused - for those reasons we will henceforth focus on the term machine learning.

Machine learning itself is a field of study aimed at building machines with the ability of learning from given data to find patterns or otherwise gain an understanding of said data. Various approaches could be based on statistics, clustering, transformations or on deep learning, but in nearly all cases, they are heavily dependent on the given data (so-called datasets described below) and successful learning on one problem does not ensure satisfactory results on another input. A common division of machine learning methods is according to the learning process into supervised and unsupervised methods - depending on whether the inputs have a known (labelled) content at the beginning of the learning process or not. If the input data could be labelled, the supervised approach requires more work in the preparation stage, but in comparison with the unsupervised methods, it usually brings more satisfactory results. Nevertheless, input data might not always allow labelling and the unsupervised learning approach must be used in this case \cite{russell2009edition}.

For the purposes of this paper, we will divide those methods into three groups - conventional classifiers, deep learning classifiers and object detectors, which are all described in more detail below. The main difference between the conventional and deep learning classifiers is in the way the features for classification are selected – they are hand-picked by the programmer (e. g. the observed object size, colour, or various other descriptors) in the conventional classifier case, or they are obtained by the model as part of the learning process in the deep classifier case.

An advantage of the deep learning approach is that the model itself finds the most suitable features to describe the object, which could be often problematic to get by hand. On the other hand, those features might sometimes not be the best representatives (for example fragments of background or general biases in the data) and a larger amount of data is typically needed for the learning process. Object detectors are usually based on the deep learning approach, and they perform both classification and localization of the searched object \cite{horak2019deep}.

\subsection{Conventional computer vision and machine learning techniques}

Conventional computer vision (CV) tasks are commonly divided into image filtering, image segmentation and image recognition. The classical machine learning methods could be categorized as statistical methods, various feature space transformations, support vector-based classifiers and simple neural networks \cite{8675097}. These methods do not use deep learning, but rely on analytical and statistical approaches, or classical signal processing. This leads to higher accuracy, shorter inference times and better analytical description of these algorithms, but on the other hand, they are less universal than the CNN-based ones and they offer no robust solution for some very complex problems (written text detection, speech recognition, etc.). Nevertheless, better transparency and explainability of these methods remains their great advantage in comparison to the CNNs, which are more difficult and complex to describe. It must be noted that in the field of explainable AI (also known as XAI) advances are being made to mitigate this problem \cite{9050829}.

\begin{figure}[htb]
    \centering
 	\includegraphics[width=0.9\textwidth]{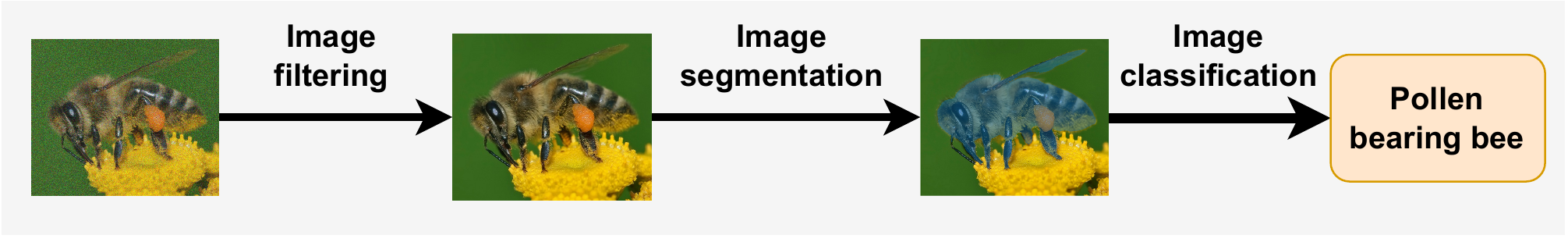}
 	\caption{Example of the conventional computer vision detection pipeline}
 	\label{Fig:CVPip}
\end{figure}

In the conventional approach (example shown in Fig. \ref{Fig:CVPip}), the common task like object detection or object classification is usually divided into several individual stages. Every stage is tuned by hand to obtain the data for the next stage. The first stage is commonly some computer vision algorithm in the role of a feature extractor. The second stage is a suitable machine learning algorithm tuned for this extracted data to complete the intended task. This approach is best suited for industry-like environments, where controlled surroundings can be used to simplify the task. The main benefit is lower computational complexity.

\subsection{CNN based classifiers}

A very specialized type of supervised Machine Learning techniques is the class of Deep Learning, and more specifically, the Deep Convolutional Neural Networks. Convolutional Neural Networks, first made practical in the 1980s in \cite{6795724}, showed a lot of promise but were prevented from further proliferation into more complicated tasks by their computationally demanding nature. Due to the increase in availability of processing power in the 1990s and 2000s deeper and deeper networks could be implemented. Since the introduction of AlexNet in 2012, which decisively won the 2012 ImageNet contest, CNNs have become the de-facto state-of-the-art in many computer vision areas \cite{10.1145/3065386}.

Convolutional neural networks have an advantage in the great complexity and huge number of parameters. This results in the ability to make all the necessary steps needed to classify an object in one step without the need for almost any pre-processing from computer vision algorithms. Convolution layers are automatically focused on sub-parts of the image, and if key objects are found, the object is classified into a learned class. All parameters, including convolution masks, are set automatically, but learning from the beginning can be very time-consuming. Therefore, networks taught for a general problems are used as a template and are only re-trained for a specific task.

\begin{figure}[htb]
    \centering
 	\includegraphics[width=0.9\textwidth]{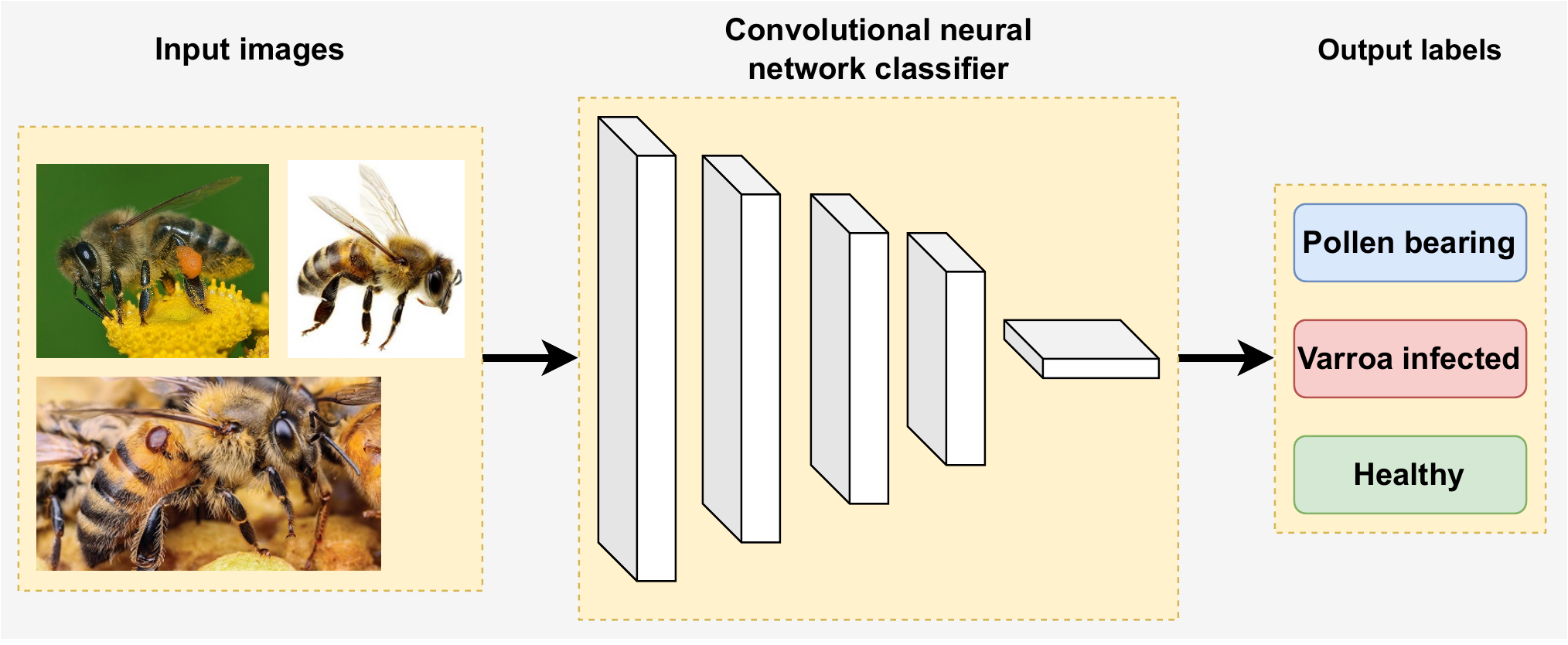}
 	\caption{Example of the convolutional neural network classifier pipeline}
 	\label{Fig:CNNPip}
\end{figure}

The main disadvantage of the deep learning approach are its high computational requirements for the real-time detection. However, in recent years, a large number of computing units adapted for mobile applications such as NVIDIA Jetson, or various accelerators applicable to RaspberryPi have emerged that are suitable for online classification. The training itself, which could be extremely computationally demanding, could be performed externally using specialized hardware. The description above is obviously very brief and simplified, because the neural network techniques are a very complicated and fast-evolving field as described for example in \cite{horak2019deep}. An example of this approach is shown in Fig. \ref{Fig:CNNPip}.

\subsection{CNN based object detectors}

Convolutional neural networks are not only used for classification, but are also applied in object detection in real life environments. A convolutional neural network automatically generalizes the key features of the searched object and is able to distinguish this said object from the background.

Convolutional neural networks were introduced in the previous section. As a reminder, convolution neural networks are machine learning algorithms, which can classify an image in one of the categories. These algorithms are called classifiers. On the other side, an object detector can detect, classify and locate the instance in the image (rather than that it simply is in the image). Further, object detectors can identify multiple objects in a single image. %An example of a difference is shown in Fig. \ref{Fig:ClassVsDetect}.

\begin{figure}[htb]
    \centering
 	\includegraphics[width=0.9\textwidth]{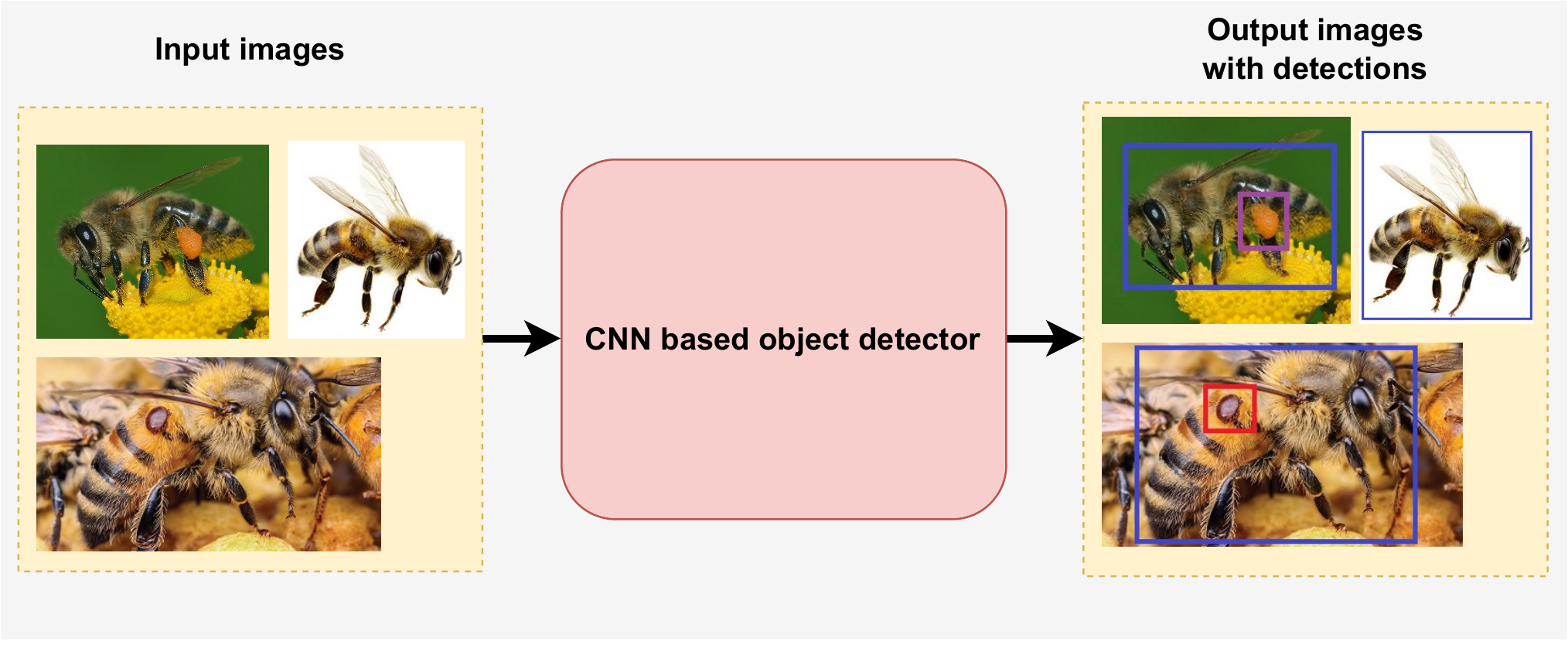}
 	\caption{Example of the object detector pipeline}
 	\label{Fig:ODPip}
\end{figure}

The object detector model (example shown in Fig. \ref{Fig:ODPip}) structure is similar to that of a CNN classifier, however it is complemented by a way of looking through the analysed image for instances of searched for classes. One way of achieving this is to enrich the CNN classifier with grid and anchor boxes. Anchor boxes are hyperparameters and make it possible to find more objects in one image. A feature image from a convolution layer is divided into a grid and on every grid cell, we try to find the anchor boxes. When the confidence is higher than a threshold, the box is returned as a detection.

\subsection{Available bee datasets}

As mentioned above, machine learning techniques require a large amount of input data called datasets. For supervised learning such data must be annotated (labelled) by an expert human annotator before the learning and testing stage of the selected method. Automated or semi-automated tools for annotating datasets exist, however even when utilized, it is a costly and time consuming process. Preparation of a proper dataset is crucial for the quality of the results - for conventional machine learning algorithms, it should contain a similar amount of data for all examined classes in the experiment in the required quality. It is often a complication to fulfil this requirement, for example because instances of some classes could be much rarer than others.

For these reasons, data augmentation (techniques used to increase the dataset size using only the existing samples) is often employed in order to increase the amount of input data. For working with images we use techniques such as rotation, blurring, scaling, various colour filters, or approaches involving machine learning, which could help achieve better learning results and a greater robustness of the selected model. On the other hand, a sufficient amount of information about the observed data has to be already present in the original dataset \cite{devries2017improved}. Important factor may also be the background of the images. Image background can increase or decrease the difficulty of classification/detection. If we could control the capture scene, we may get better results.

Several bee datasets are available, containing various types of related data (images, sound, hive weight), but each of them has been compiled for a different application and thus each is tailored for it. Type of data and the context of its labels determines the field of usability of every dataset. Some contain object position in an image, classification from a list of classes, and/or some general information such as location of data acquisition. For an apiarist additional information may also be useful, such as the bee subspecies, beehive type or acquisition time and location. Sometimes it is difficult to obtain this information. For datasets covered by this article we have prepared a summary in Table \ref{Tab:datasets}.

\begin{table}[width=.9\linewidth,cols=5,pos=h]
    \caption{Datasets with additional information about bee kind, beehive type and acquisition time and location. Rows are sorted by the amount of available information.}
    \label{Tab:datasets}
    \begin{tabular*}{\tblwidth}{@{} p{2,6cm}p{2,4cm}p{2cm}p{3cm}p{3cm}LLLLL@{} }
        \toprule\toprule
        \textbf{Citation} & \textbf{Bee subspecies} & \textbf{Hive} & \textbf{Time} & \textbf{Location}\\
        \midrule\midrule
        \cite{BeePi} & Carniolan, Italian & Langstroth & 4/2017 - 9/2017, 5/5/2018 - 6/2018,  6/2018 - 7/2018, 5/2018 - 11/2018, 4/2018 - 7/2019 & North Logan, Logan (UT), USA\\
        \midrule
        \cite{BUT1} & Carniolan & Langstroth & 9/2021 & Darkovice, Czechia\\
        \midrule
        \cite{BUT2} & Carniolan & Langstroth & 6/2022 & Darkovice, Czechia\\
        \midrule
        \cite{Yang2018} & Carniolan, Italian, Russian & Not specified & 2/7/2018 - 8/9/2018 & Saratoga (CA), Des Moines (IA), Alvin (TX), Athens (GA), USA \\
        \midrule
        \cite{8354145DS} & Not specified & Not specified & 7/2017 & Gurabo, Puerto Rico\\
        \midrule
        \cite{BeeAlarmed} & Not specified & Langstroth & 2019 - 2020 & Not specified\\
        \midrule
        \cite{9286673DS} & Not specified & Laboratory & Not specified & Wien, Austria\\
        \midrule
        \cite{beemon} & Not specified & Not specified & 2019 - 2023 & Not specified\\
        \midrule
        \cite{s21082764} & Not specified & Not specified & Not specified & Not specified\\
        \midrule
        \cite{Rey2020} & Not specified & Not specified & Not specified & Not specified\\
        \midrule
        \cite{YOLO} & Not specified & Not specified & Not specified & Not specified\\
        \bottomrule
        \bottomrule
    \end{tabular*}
\end{table}

The following sections discuss which datasets are suitable to solve the application tasks in the categories of the pollen detection, Varroa mite detection, bee traffic monitoring and general bee inspection. Each application field has its specific requirements on input data, but some of the datasets could be used in multiple fields. 

\subsubsection{Datasets for pollen detection}

In this application a visibility of the carried pollen is critical. Dataset \cite{BeeAlarmed} contains high resolution images capturing bees in the dominant part of the image with pollen and has around 7500 samples. The second dataset \cite{8354145DS} also contains high resolution images of the pollen-bearing bees and is presented in \cite{8354145}, but has less samples, around 800. The third suitable dataset \cite{Yang2018} contains a high number (around 5000 images) of bees with extensive amount of additional information about their health-state, whether they are pollen carrying, or subspecies, but in a low resolution. The fourth dataset that could be used for this type of application is our previous work \cite{s21082764} and is created from images available online and it is available upon request. Datasets \cite{BUT1} and \cite{BUT2} also seem suitable, but do not have pollen-bearing state labels.

\subsubsection{Datasets for Varroa mite detection}

For this task it is necessary for the Varroa mite to be visible (the mite may attach to nearly any part of the bee). The authors of \cite{9286673} published their dataset for the Varroa mite detection in \cite{9286673DS}, it contains short videos of healthy and infected bees in high resolution. The dataset has around 13500 samples that were collected in a laboratory environment. Dataset \cite{BeeAlarmed} is a public dataset and contains around 7500 samples annotated for the Varroa mite infection status. In the article \cite{s21082764}, we used the dataset mentioned above in the paragraph about Varroa mite detection. 

\subsubsection{Datasets for bee traffic monitoring}

 In a bee traffic monitoring the actual bee visibility and context of its movement is the main dataset requirement. The \cite{BUT1} is a dataset with 1000 samples in five classes useful for bee traffic monitoring. The authors of \cite{YOLO} created two videos and shared them on Youtube, but they are no longer public. Last but not least the author of \cite{BeePi} created several datasets for traffic monitoring, but only some of them were published.

\subsubsection{Datasets for general bee inspection}

The \cite{BUT2} is a dataset with multimodal data: image, sound, humidity, CO$_2$ concentration and daylight level. The authors of \cite{beemon} created a website for streaming data from an IoT-based beehive monitoring system. Unfortunately the website does not work anymore. But from \cite{YOLO_BR} it is implied that that the dataset contained labels about different bee roles. Dataset \cite{BeeAlarmed} has included information about wasp and bee differences.

The author of \cite{BeePi} created two datasets. The first collected 43 different publicly available weather and climate variables. He connected them with acquired data from the beehive. In the second dataset 7 quantities were measured on a beehive: temperature, atmospheric pressure, relative humidity, wind speed and direction, rainfall, solar irradiance, electro-magnetic frequencies, radio frequencies and electric fields. The second datasets is not published yet.

%There are several datasets containing bee images. Dataset \cite{8354145DS} contains high-resolution images of the pollen-bearing bees and is presented in \cite{8354145}. The authors of \cite{9286673} made available their dataset for the Varroa mite detection at \cite{9286673DS}, which contains short videos of healthy and infected bees in high resolution. Dataset \cite{Yang2018} contains a high number of bees with extensive amount of additional information about their health-state, whether they are pollen carrying, or subspecies, but in a low resolution, and \cite{Rey2020} is designed for distinguishing between bees, wasps and other insect species. The authors of \cite{s21082764} use a dataset created from images available online, but it was not made public.

\section{Fields of application}\label{Sect:Applic}

For the purposes of this paper, we decided to split the existing application areas into pollen detection, Varroa mite monitoring, bee traffic monitoring and bee inspection in general.

\begin{figure}[htb]
    \centering
 	\includegraphics[width=0.9\textwidth]{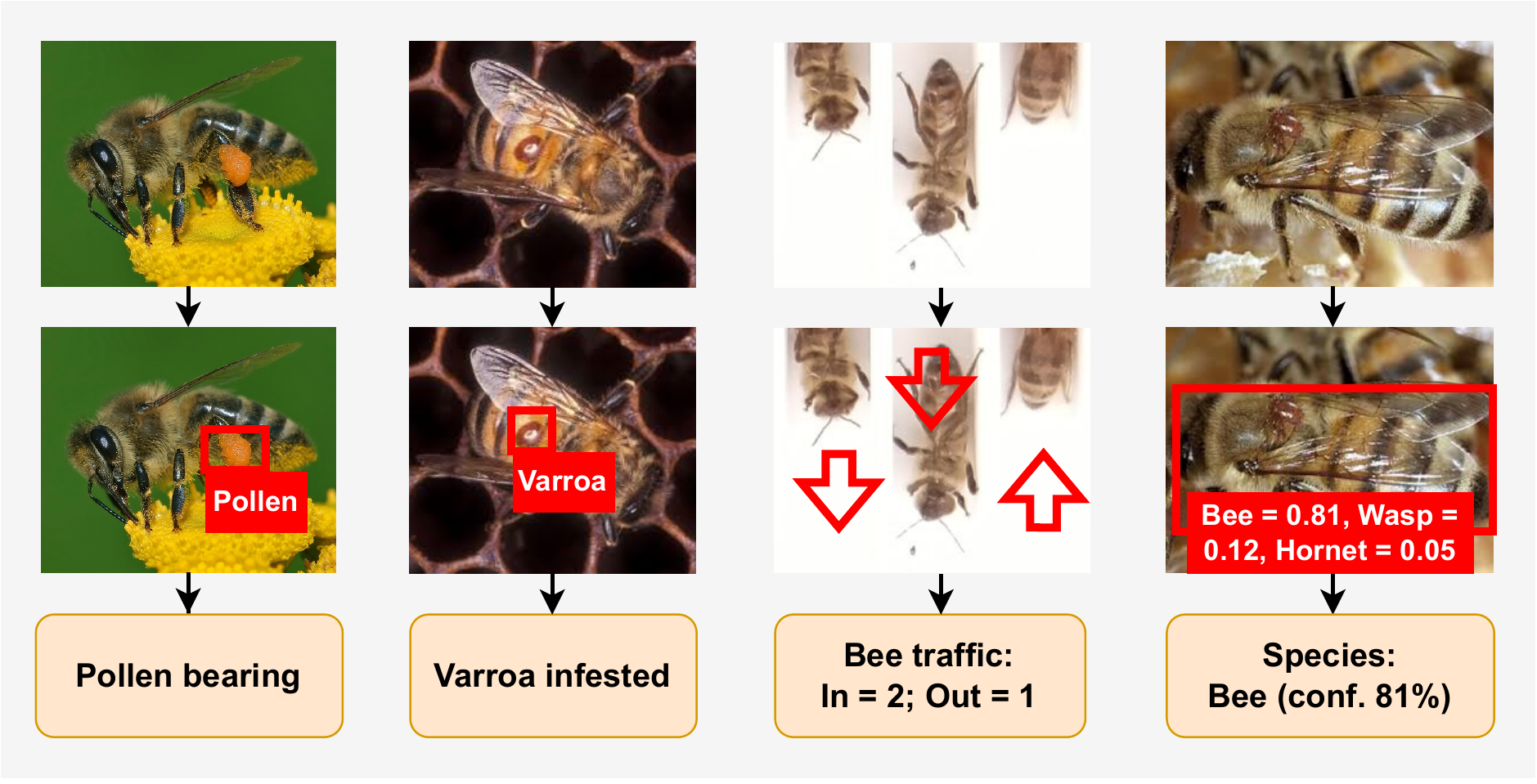}
 	\caption{Examples of Pollen detection, Varroa monitoring, Bee traffic monitoring and general bee inspection (example of intruder species identification) respectively}
 	\label{Fig:Apps}
\end{figure}

The first three applications are the most common research fields, which we found during this survey. In the general bee inspection section, we put all remaining papers which could not be categorized into the groups above. Illustration of all four investigated applications is shown at Fig. \ref{Fig:Apps}.

\subsection{Pollen detection applications}

In this pollen detection field, the research focuses on the detection of pollen-bearing bees at the beehive entrance, or in the field. It is possible to analyse a count the total of pollen-bearing bees, the size of the pollen stockpiles, or with a sufficiently large and well-annotated dataset even the pollen species. This might be useful to estimate the amount of pollen supplies in the beehive, pollination location, or the number of active flying bees. An efficient example of this application is the paper \cite{8354145}.

\subsubsection{Conventional techniques based pollen detection}

A study using classical computer vision techniques is presented in \cite{babic2016pollen}, where the authors recognize the pollen-bearing bees with the classical computer vision methods. They use a compact portable system based on the RaspberryPi 2 embedded computer, which is placed at the beehive’s entrance. The study contains its own dataset in a low resolution. The method has achieved a test accuracy around 88\%.

The authors of the paper \cite{Yang2017} employ conventional computer vision techniques to detect, track and classify bees into pollen sac carrying, and non-carrying from a video recording of a beehive’s entrance. A Gaussian mixture model in conjunction with colour thresholding is used to separate the individual bees from a fixed background, a combination of Kalman filter and Hough transform are used to track the bees across video frames. Ellipsis fitting and subtraction is then used on the detected bee blobs for classification of the sacs attached to the bee’s body (represented by the ellipse). Authors note a relatively low accuracy (which could not be specified due to a strange methodology), but conclude it provides useful information about the hive’s pollen storage situation.

In \cite{8345546}, the authors employ conventional image processing methods to detect and classify bees according to their pollen bearing status at the hive’s entrance. A colour based segmentation of foreground from controlled background is implemented in the CIE LAB colour space. For classifications the segmented subimages are resized and features are extracted using SIFT and VLAD (Vector of Locally Aggregated Descriptors), further classification is SVM based.

\subsubsection{CNN classifiers based pollen detection}

An application of CNNs similar to the \cite{babic2016pollen} is shown in \cite{8354145}. The authors created a portable system for in-field measurements of the beehive’s entrance, which captures the entering bees on a defined background. The data is processed by conventional and CNN-based classifiers in order to recognize the pollen-bearing bees with the test accuracy over 95\%.  The authors also created and published their own dataset.

The author of the study \cite{8592464} investigated the use of a custom CNN classifier for classification of pollen bearing bees in images taken on the beehive’s entrance. The intention was to assess the feasibility of the use of CNN as a classifier to be implemented in the future embedded solution on a FPGA. The resulting CNN proved to classify with 94\% accuracy on the author’s own dataset, and was judged to be fit for implementation into the hardware along with a suitable RoI (Region of Interest) selector.

\subsubsection{Object detection based pollen detection}

The paper \cite{8961011} is building on the previous paper \cite{Yang2017} by the same group of authors. Using the same setup and modified detection algorithm, the classification is handled by a Faster RCNN network trained to detect individual pollen sacs, performance is much improved compared to the previous paper.

\subsection{Varroa mite detection applications}

Considering the Varroa mite detection application field, the researchers aim to detect the presence of this parasitic mite usually with image-based systems. Because this parasite causes great bee colony losses worldwide, development and deployment of this application could also lead to a great economic and bee population savings, better-planned treatment, and continuous inspection of the bee colony’s health state. Some suggestions also exist of a system capable of destroying the Varroa mite directly on the body of the infested bee in \cite{7850001}, the mentioned system as is is immature and not ready for deployment, but perhaps some future effort on the topic could be made practical. Nevertheless, a problem of the visual-based systems is that most of the Varroa mites occur on the bee brood, or that they are hidden between the bee’s abdominal segments and poorly visible. Therefore, an infestation level estimation methodology like the one presented in \cite{BJERGE2019104898} has to be developed and extensively tested to allow a better use of this application.

\subsubsection{Conventional techniques based Varroa mite detection}

An interesting study is described in \cite{elizondo2013video}. The authors capture a video of an infected bee brood, where they track the Varroa mite location, its movement and behaviour. The experiment was performed under laboratory conditions and the authors use conventional image processing methods to subtract the background and to localize the Varroa mite. This experiment might be further improved on, by using state-of-the-art object detection methods and we believe that it might be useful for future Varroa mite behaviour analysis.

A perfect illustration of the classical computer vision approach used for Varroa mite detection is presented in the article \cite{schurischuster2016sensor} and the following \cite{schurischuster2018}, where the authors present two experimental setups to capture video of live bees in laboratory conditions. In the first setup, the bees are separated by narrow tunnels with contrasting backgrounds, which simplifies the segmentation and analysis. The device is equipped with artificial lighting to create proper recording conditions. No details are given about the light source used in the text but from the figures it seems that white LED strips producing diffused light are used. The second setup captures the bees in the open with a camera placed above the measured surface, which speeds up the measurements process, but complicates further processing. The classification stage is designed to differentiate between healthy bees without Varroa mites and infected bees with Varroa mites. Presented work compares various feature sets with three different classification models (Naive Bayes, SVM and Random Forest) resulting in accuracy and F1-score over 0.8 for some combinations.

The authors of \cite{BAUER2018311} use an IR camera to detect the temperature increase of the infected brood cells. They prove a slight temperature increase of the infected cells, the cause of which is still unclear. The measurements were performed in laboratory conditions and in the current form, they would be quite impractical for in-field measurements. The article nevertheless shows a certain potential for the thermal measurements approach.

Another usage of conventional machine learning methods is presented as a part of \cite{BJERGE2019104898}, where the best light wavelength combination was sought by Linear Discriminant Analysis (LDA) to train and test a classifier looking for the resulting three-dimensional vector of wavelength components that is the most useful for separating bees and mites. The best results for bee mite separation are for light wavelengths above 700 nm. The chosen combination was 470-630-780 nm. In addition, the use of spectra above 700 nm will not impair the comfort of bees, because the bee’s eye is not sensitive to higher wavelengths \cite{HempeldeIbarra2014}.

The author of \cite{konig2020varroacounter} introduces his automatic system for the Varroa mite detection on a bee debris plate as a part of his more complex bee health monitoring system presented in \cite{konig2019indusbee}. The author proposes a two step system, where he uses a classical CV-based blob analysis to detect candidate regions with a fallen Varroa mite followed by a k-NN classifier to detect Varroa mites within those regions. This application is similar to the paper \cite{9897809}.

Another Varroa mite monitoring system is presented in \cite{9964541}. Authors present a prototype of a Raspberry Pi based system, which is built-in to the bee hive frame with a camera, internal illumination and liquid lens allowing focus on very close distance of the bee space. The whole device should be placed inside the beehive close to the cells with a fresh brood and it is supposed to detect Varroa mites on the youngest bees. The authors suggest a possibility of long-term monitoring and emphasize a low disturbance of such a system to the beehive and they mention that a robust autofocus algorithm of this system still remains a challenge. Presented device is currently under testing.

In the study \cite{jimaging9070144}, the authors use Legendre-Fourier moments within various colour spaces as RGB, HSV and YCbCr to detect Varroa mite infected bees on the \cite{9286673DS} dataset. The authors compute moment based descriptors separately for all channels of the selected colour space and they connect all three descriptors to a single feature vector. Classification is two-staged, where the authors first classify the bee orientation (dorsal, or ventral) and then the Varroa mite presence on each orientation separately. This method reached the maximal F1-score of 0.96, which surpasses the YOLOv5 F1-score of 0.86 and which is comparable to the semantic segmentation using the DeepLabV3 framework.

\subsubsection{CNN classifiers based Varroa mite detection}

The research presented in \cite{7850001} suggests a system for visual Varroa mite detection and its termination using a swept laser beam controlled by a RaspberryPi 3 embedded computer. In future development, the authors plan to place a camera with the above-mentioned laser system above the beehive’s entrance and use a CNN classifier for the Varroa mite recognition followed by its localization and precise laser beam hit. Although the experiment looks promising, it would be very difficult to achieve the planned results with the proposed equipment and not to harm the bee by an inaccurate hit. Furthermore, much development will be required if a practical mature product is to be developed.

Probably one of the most complex studies is the previously mentioned \cite{BJERGE2019104898}. The authors describe a device suitable for in-field measurements, which is based on the findings presented in \cite{schurischuster2018}. The device uses narrow tunnels to separate the bees together with multispectral illumination and a camera located beneath the tunnels. The entire device is designed to be located at the entrance of the beehive. For the individual bee detection, the authors use conventional computer vision techniques with custom CNN for the Varroa mite detection and localization. The authors also suggest a methodology (ILE) for estimation of the infestation level of the whole beehive without a need to measure all bees based on the detected Varroa mites and counted bees.

One example of a common use of convolutional neural networks in the role of a classifier is presented in \cite{9286673}. The authors utilize a semantic segmentation approach using DeepLab3 and classification methods based on AlexNet and ResNet in order to separate healthy and Varroa mite infected bees. Because no exact mite images were presented, the authors created a new dataset \cite{9286673DS} with 13500 annotated images with healthy and mite infected bees. This paper found good results with ResNet101 architecture pre-trained on ImageNet, or AlexNet architecture trained from scratch, but the best performing models use DeepLabV3 to identify pixels belonging to the parasite and therefore perform semantic segmentation with two variants of ResNet architecture as encoding models. Per class accuracy of minimum 90\% was reached during the experiments and the results prove the potential of the semantic segmentation approach.

The authors of \cite{app112211078} propose an embedded system for Varroa mite monitoring based on Raspberry Pi 4 with the Coral TPU accelerator. This system uses a camera placed in front of a bee hive, which captures the bee traffic and a two stage system for the Varroa mite detection with the detection of the individual bees as a first step and with the detection of the Varroa mites as the second one. Both models were built with the Google AutoML system. The proposed setup reaches an F1-score around 0.8 in both bee and Varroa mite detection applications. The whole system runs on the Amazon Web Service (AWS) and is complemented by automated notifications of the beekeeper.

In \cite{9897809} the authors propose a method to determine the degree of beehive varroa mite infestation by detecting and counting varroa mites in the hive’s fallout. As opposed to some other methods the presented approach does not require much specialized equipment save for a smartphone to take an image of the fallout. The image then goes through thresholding RoI preselection, and CNN based classification using a shallow CNN called VarroaNet (variants VarroaNet-0.1 and VarroaNet-0.05 with 100k and 50k parameters respectively). The presented results of mean infestation level accuracy of 96.0\% for autumn and 93.8\% for winter are promising and given the architecture of the system computational requirements are minimized, however the authors themselves admit that the RoI preselection fails in certain situations, such as occlusions.

In \cite{9988275}, the authors utilize a custom CNN classifier to detect Varroa mite infestation on the bee images with the claimed F1-score of 0.99. To solve the problem of insufficient training data, authors compare the effect of several data augmentation techniques including Contrast Limited Adaptive Histogram Equalization (CLAHE), geometrical transformations and DCGAN generative architecture. Best results were achieved with the CLAHE augmentation.

\subsubsection{Object detection based Varroa mite detection}

An object detector approach for Varroa mite detection is shown in the paper \cite{s21082764}, where the authors use CNN-based object detection techniques SSD and YOLO to classify and localize the Varroa mite in a single step. The authors use a custom dataset with bees in various locations to train the networks in order to classify bees and Varroa mites. The best results were obtained in the classification of infected bees using the YOLO object detector. The main advantage of this approach, compared to the previous ones, is the potential of real-time measurement and analysis due to short inference times of the selected algorithm. Nevertheless, a practical evaluation of this experiment has not been performed yet.

The authors of \cite{signals3030030} use a similar approach as in \cite{9964541}, where they use one camera monitoring the brood in the opposite beehive frame and the second one, which captures the tops of the frames. Frame holding the cameras also contains custom electronics, which facilitate sending data through WiFi in the online mode. The whole system is powered by an external battery and a solar cell. The authors use a Fast-RCNN for the individual bee detection and the Hough-transform for the Varroa mite detection with a bee detection accuracy of 88\% and a mite detection accuracy of 77\%. A disadvantage is that the cameras have to be placed at a much higher distance than is a bee space, which might lead to brood cooling.

In the paper \cite{10104935} the authors primarily use sound analysis and Mel spectrogram on their own sound dataset to distinguish strong and weak bee colonies. To have a comparison with the other techniques, they use SSD, various YOLO and DETR object detectors together with the dataset \cite{9286673DS} containing healthy and Varroa mite infested bees. Using the object detection techniques, authors reached the best mAP score of 0.97 using the YOLOv5s model.

\subsection{Bee traffic monitoring applications}

Another field, which could be improved with the computer vision techniques is the bee traffic monitoring. Although there exist various bee counting systems as described in \cite{https://doi.org/10.1111/aab.12727}, camera based systems could be easily supplemented by other functionalities such as pollen detection, or Varroa mite monitoring as described above. These systems could be also implemented in a non-intrusive way, which is advantageous in comparison with tagging the bees with RFID or optical pattern tags. The described systems might be very useful in the case of bee loss early warning systems or for general bee behaviour analysis.

\subsubsection{Conventional techniques based bee traffic monitoring}

The study \cite{CHEN2012100} presents a visual-based system for monitoring incoming and outgoing bee traffic. The authors created a device with an infrared camera and illumination, which was placed in front of a beehive and allowed the bees to enter through narrow passages. A small sample of the bees was tagged by a visual identifier and their activity was recognized using Hough transformation and SVM classifier. Despite the impractical and demanding tagging approach, the device design is sound and might be an inspiration for other authors.

In the paper \cite{Chiron2013} a 3D tracking approach of tracking bees entering and exiting the hive is explored. The authors outline challenges of 3D tracking of small, fast-moving and interacting objects, and build a data acquisition system based on stereo camera setup. A hybrid tracking method is proposed, and integrated with a Kalman filter based tracking using  Global Nearest Neighbor or Multiple Hypothesis Tracking to associate unknown datapoints to known tracks. The approach, while not mature for deployment, seems to be promising and offers a new perspective of 3D tracking into the established 2D tracking paradigm.

In the series of the papers \cite{app10062042}, \cite{BeePi}, \cite{kulyukin2021beepiv}, \cite{kulyukin2022integration} and \cite{s23156791}, a research team presents their system for bee monitoring, collected dataset and several methods for bee traffic monitoring. In the first paper \cite{app10062042} the authors build on the data acquired using data gathering device later presented in \cite{BeePi} and present an image velocimetry based method for bidirectional bee traffic monitoring. They compare the achieved results with the method presented in the previous paper. The proposed method is time demanding and the authors also present a network of 6 RasberryPi3 embedded computers to process the collected data within the 24 hours capture period. The study misses a more detailed analysis of the algorithm accuracy.

The method described in the second paper \cite{BeePi} is based on a combination of Gaussian Motion Detection (MOG2) algorithm followed by a classifier, which tohether recognise if the detected region corresponds to a honey bee. Nevertheless, this algorithm is not able to recognize the motion direction.

In the third paper \cite{kulyukin2021beepiv}, the authors present a novel BeePIV algorithm for bee movement monitoring. Applying particle image velocimetry in video of incoming and leaving bees in front of a hive entrance a classification of inbound, outbound and lateral bee movement is performed, and bees are counted. After monitoring multiple hives for 7 months interesting patterns emerged while monitoring healthy and failing hives. While a healthy hive traffic showed periods of inbound and outbound traffic, a failing hive displayed no such variation. The authors believe the system might aid in timely hive state diagnosis and could help prevent hive failure.

In the fourth paper \cite{kulyukin2022integration}, the authors further investigate the relationship between the often measured hive weight, and bee traffic using their previously published \cite{BeePi} monitoring equipment and methods. Through statistical methods on a large dataset the authors have concluded that hive weight and bee traffic are correlated over longer periods of time (while over shorter periods of time the correlation could not be conclusively proven). The authors note that certain hives had different correlations of different bee traffic indicators, and speculate on the possible causes. The paper adds to the state-of-the-art by providing a link between the often examined measures.

\subsubsection{CNN classifier based bee traffic monitoring}

An embedded device is presented in \cite{app9183743}, where the authors use a RaspberryPi 3 embedded computer with a camera module to capture bee traffic on a beehive entrance. The camera is placed above the beehive entrance and captures the bee traffic in periodical intervals. From the collected data, the motion regions are detected and processed by a CNN based classifier in order to recognize and count individual bees. In the future, this method is to be used for bee traffic estimation. The authors made their datasets publicly available.

\subsubsection{Object detection based bee trafic monitoring}

In the paper \cite{10.1371/journal.pone.0239504} the authors propose a system for insect monitoring in their foraging environment. The proposed HyDaT (Hybrid Detection and Tracking) algorithm tracks individual insects offline, one by one in a video sequence for the purpose of recording actions taken by the insects in the foraging environment. YOLO is used as the deep learning image object detector, and a KNN-based segmentation for background separation. A “predict and detect” approach is used to associate the detection and the insect of interest. The authors see the major contribution to the state-of-the-art in the application of modern computer vision approaches to pollinator tracking in complex environments, and showing that computationally inexpensive algorithms can be successfully used in detection and tracking tasks allowing low-power devices - such as Raspberry Pi - to run these applications.

A complex study \cite{NGO2021106239} presents a real-time system based on NVIDIA Jetson complemented by a Raspberry Pi 3 based environmental sensing module, which aims to detect pollen-bearing bees and to research the relationship between the environmental factors and pollen-bearing bee rate. For this purpose, authors use YOLOv3-tiny object detection model to detect bees followed by a Kalman filter used for their tracking. The authors also use tracked bee paths to determine if the bees are incoming or leaving and they can process the data with a frame rate of 25 FPS. Using the bee traffic and pollen-bearing data, authors compute several parameters describing bee colony activity. The overall F1-score for pollen and non-pollen bearing bee recognition is 0.94.

Authors of the another study \cite{ryu2021honeybee} aim on counting bees entering and leaving the beehive using the YOLOv4 object detection. Proposed system uses a special beehive entrance consisting of parallel tunnels from acrylic glass equipped with a special valve, which ensures that bees move only in one direction. This entrance is captured by a camera placed above the tunnels. Individual bees are recognized using the YOLO model and tracked using the DeepSort model. Detection accuracy of the proposed approach is 99.5\%.

In the study \cite{YOLO}, the authors used their own videos of the bee traffic in front of a hive to create a bee tracking framework. It consists of a YOLOv5 object detector trained on the bees with and without the pollen used for the bee localization and StrongSort method with MobileNetV2 and OSNet networks for the bee tracking. This approach reaches a precision of 0.98 for detecting both bee classes in the detection part and ATA score of 0.8 for the tracking part. Nevertheless, ATA score significantly drops to 0.2 with an increase of the bee traffic and a higher amount of bees in the video. Inference times for single frames varied on the used hardware between 70-150ms, which should be suitable for the real-time processing.

Authors of \cite{9788643} intend to design a complex system for bee colony health state monitoring based on the automated honey bee recognition using a camera in front of an examined beehive and SSD object detector. At this moment, SSD is used only for bee counting and the camera frames are sent to the web application. Authors used their own dataset, which is not further described or made publicly available.

In the proof-of-concept studies \cite{10134563} and \cite{10134852}, the authors focus on the bee activity monitoring on the beehive's entrance. To achieve this, they use the YOLOv8m object detector for the localization and ByteTrack package for tracking of the individual bees. Detected bee paths and their speeds are recorded as heat maps and those records are further investigated in order to recognize situations such as fanning, foraging and guarding. Presented situations could be distinguished from each other, but the authors acknowledge a need of a more extensive dataset. Detection accuracy of the individual bees reaches 0.97 mAP.

The authors of the \cite{BeePi} published a follow-up study \cite{s23156791}, which aims on comparison of the accuracy and energy footprint during the training and detection stages of several YOLO models used for the bee traffic monitoring on the experimental setup presented in the original paper. The authors recommend to use YOLOv7 model, which has the smallest energy footprint and they emphasise the importance of considering also the energy consumption aspects of the models used.

\subsection{General bee inspection applications}

In the general bee inspection field, the experiments often focus on distinguishing between the bees and other insects. This could be particularly useful for example in the Asian giant hornet and other intruder’s detection, or bee anomalous appearance detection, where the results could lead to a fast response of the beekeeper and potential damage mitigation. This application field also covers the brood cell inspection field as in \cite{colin2018development}, or bee grooming behaviour investigation described in \cite{GIUFFRE2017338}, where machine learning and computer vision techniques might be very time-saving and easily used for the inspection and research purposes.

\subsubsection{Conventional techniques based general bee inspection}

Several studies focus on the visual analysis of the brood cells. One of those studies \cite{knauer2007comparison} compares several conventional classification techniques in order to recognize the brood cells uncapping and shows the possibility of such in-field measurements, even with the presence of bees. Another study \cite{ALVES2020105244} covering the same topic is described in the following section focused on the CNN-based methods.

In \cite{4359330}, the authors apply a state-of-the-art algorithm in bee tracking and behavior prediction on the problem of bee dance analysis. They use a generic model of the insect consisting of three ellipses and use it to track movements of the bee in a video sequence. Further they design and implement a Markov chain based motion model that serves as a behavior model. For a particular dance, the waggle dance, they also estimated parameters of interest describing the dance.

An interesting study using classical image processing techniques is presented in \cite{GIUFFRE2017338}, where the authors explore the bee grooming process by observing the bees covered with baking flour. For the grooming process analysis conventional computer vision methods were used, such as colour corrections and binary segmentation.

The authors of \cite{colin2018development} propose a conventional CV based method for counting brood from frame images called CombCount. The system implemented in Python OpenCV classifies and counts comb cells into categories such as uncapped, capped honey, capped brood and more, and segments the images accordingly. Not only is the count information provided, but also the distribution of the classes, allowing for further study. The authors provide data on newly formed colony brood and honey store development.

\subsubsection{CNN classifiers based general bee inspection}

In \cite{Marstaller_2019_ICCV} a cloud-based CNN model called DeepBees is introduced. The aim is to provide an integrated multi-task deep learning model for automating hive monitoring where all the modules utilize a common feature extractor. The individual tasks are represented by modules, such as the genus module (classification between honey bee, wasp, hornet and a bumblebees), the classification module (classifying pollen carrying worker bees, non-carrying worker bees, drones and dead bees), pollen module implementing an SSD based pollen detector, and a pose module estimating a bee’s pose via 32 key points. The system is now deployed and is being further developed.

The study \cite{ALVES2020105244}, similarly as \cite{knauer2007comparison}  focusing on the brood cells, extends the older one with deep learning techniques in order to detect seven possible states of the brood cell from a single image using common CNN classifiers and data augmentation techniques. The authors also provide their dataset and the analysis software.

The research presented in \cite{KAPLANBERKAYA2021101353} compares several CNN architectures on datasets presented in \cite{9286673DS}, \cite{Yang2018} and \cite{8354145DS}. The authors used the originally proposed classes of each dataset and they proved that the common CNN classifiers such as ResNet, AlexNet, or VGG are able to solve those tasks with a comparable accuracy to the original works, which used conventional classifiers or custom CNNs. The paper provides the inference times and performance analysis of all models used.

The author of \cite{10069892} compares the performance of four CNN models (VGG19, InceptionV3, MobileNet and custom CNN architecture) together with an SVM classifier fed directly by the image data. The dataset used in this paper is taken from \cite{Yang2018}. A F1-score over 0.9 with all classifiers is reached with the best performing one being the VGG19 (F1-score of 0.98).

For the sake of completeness, we mention two non-English studies, which are therefore not described in great detail. The authors of the first paper \cite{wcama} focus on distinguishing between the honeybees and other insect species. For this task, they use two publicly available datasets \cite{Yang2018}, \cite{Rey2020} and a custom CNN-based classifier. Using this architecture, they were able to reach a precision of around 94\% in the classifications task. The authors of \cite{8875886} use several custom CNN classifiers to ascertain the health status of the observed bees.

\subsubsection{Object detection based general bee inspection}

A high potential of the object detection approach for bee traffic monitoring is also shown in \cite{YOLO_BR}. Authors of this paper use YOLOv5 network and their own dataset for the detection and recognition between the bee workers and drones from the traffic at the bee hive’s entrance. The detection average precision over both classes and all test datasets was 0.92. The paper is unfortunately not published in English.

%\clearpage

\subsection{Overview table of the investigated papers, their application areas and methods used}

In this section, we summarize the papers investigated in the previous chapters, its application fields and methods used. This overview is shown in Tab. \ref{Tab:Papers}.

\begin{table}[width=.95\linewidth,cols=5,pos=htb]
    \caption{Overview of the investigated papers divided by their field of application and the method used}
    \label{Tab:Papers}
    \begin{tabular*}{\tblwidth}{@{} p{1.6cm}p{2.8cm}p{2.8cm}p{3cm}p{3cm}lllll@{} }
    \toprule\toprule
    & Pollen detection & Varroa monitoring & Bee traffic monitoring & General bee inspection \\
    \midrule\midrule
    Conventional techniques & \begin{tabular}[t]{p{2.8cm}} \cite{babic2016pollen} \\ \cite{Yang2017} \\ \cite{8345546} \end{tabular}
    
    & \begin{tabular}[t]{p{2.8cm}} \cite{elizondo2013video} \\ \cite{schurischuster2016sensor} \\ \cite{schurischuster2018} \\ \cite{BAUER2018311} \\ \cite{BJERGE2019104898} \\ \cite{konig2020varroacounter} \\ \cite{9964541} \\ \cite{jimaging9070144} \end{tabular}
    
    & \begin{tabular}[t]{p{2.8cm}} \cite{CHEN2012100} \\ \cite{Chiron2013} \\ \cite{app10062042} \\ \cite{BeePi} \\ \cite{kulyukin2021beepiv} \\ \cite{kulyukin2022integration} \end{tabular}
    
    & \begin{tabular}[t]{p{2.8cm}} \cite{knauer2007comparison} \\ \cite{4359330} \\ \cite{GIUFFRE2017338} \\ \cite{colin2018development} \end{tabular}          \\
    
    \midrule
    CNN based classifiers   & \begin{tabular}[t]{p{2.8cm}} \cite{8354145} \\ \cite{8592464} \end{tabular}
    
    & \begin{tabular}[t]{p{2.8cm}} \cite{7850001} \\ \cite{BJERGE2019104898} \\ \cite{9286673} \\ \cite{app112211078} \\ \cite{9897809} \\ \cite{9988275} \end{tabular}
    
    & \begin{tabular}[t]{p{2.8cm}} \cite{app9183743} \end{tabular}   
    
    & \begin{tabular}[t]{p{2.8cm}} \cite{Marstaller_2019_ICCV} \\ \cite{ALVES2020105244} \\ \cite{KAPLANBERKAYA2021101353} \\ \cite{10069892} \\ \cite{wcama} \\ \cite{8875886} \end{tabular} \\
    
    \midrule
    Object detection & \begin{tabular}[t]{p{2.8cm}}  \cite{8961011} \end{tabular}
    
    & \begin{tabular}[t]{p{2.8cm}} \cite{s21082764} \\ \cite{signals3030030} \\ \cite{10104935} \end{tabular} 
    
    & \begin{tabular}[t]{p{2.8cm}} \cite{10.1371/journal.pone.0239504} \\ \cite{NGO2021106239} \\ \cite{ryu2021honeybee} \\ \cite{YOLO} \\ \cite{9788643} \\ \cite{10134563} \\ \cite{10134852} \\ \cite{s23156791} \end{tabular}
    
    & \begin{tabular}[t]{p{2.8cm}} \cite{YOLO_BR}  \end{tabular}       \\
    
    \bottomrule\bottomrule
    \end{tabular*}
\end{table}

%\clearpage

\section{Discussion}\label{Sect:Disc}

%In the beginning of discussion section, we summarize the trends in the automated beehive monitoring and the popular methods at the time of writing. We also discuss possible outlooks of the future research and its trends.

%\clearpage

%\subsection{Observed trends in the automated beehive monitoring}

To describe the observed trends in the automated beehive monitoring, we use a division of the existing papers relating to the method used (conventional computer vision approach, CNN based classification and object detection) together with the application area (pollen detection, Varroa mite detection, bee traffic monitoring and bee inspection in general). We also analyzed the number of investigated papers (50 papers in total) during the last ten years and the prevalent technique used to solve the problem.

\begin{figure}[htb]
    \centering
 	\includegraphics[width=0.8\textwidth]{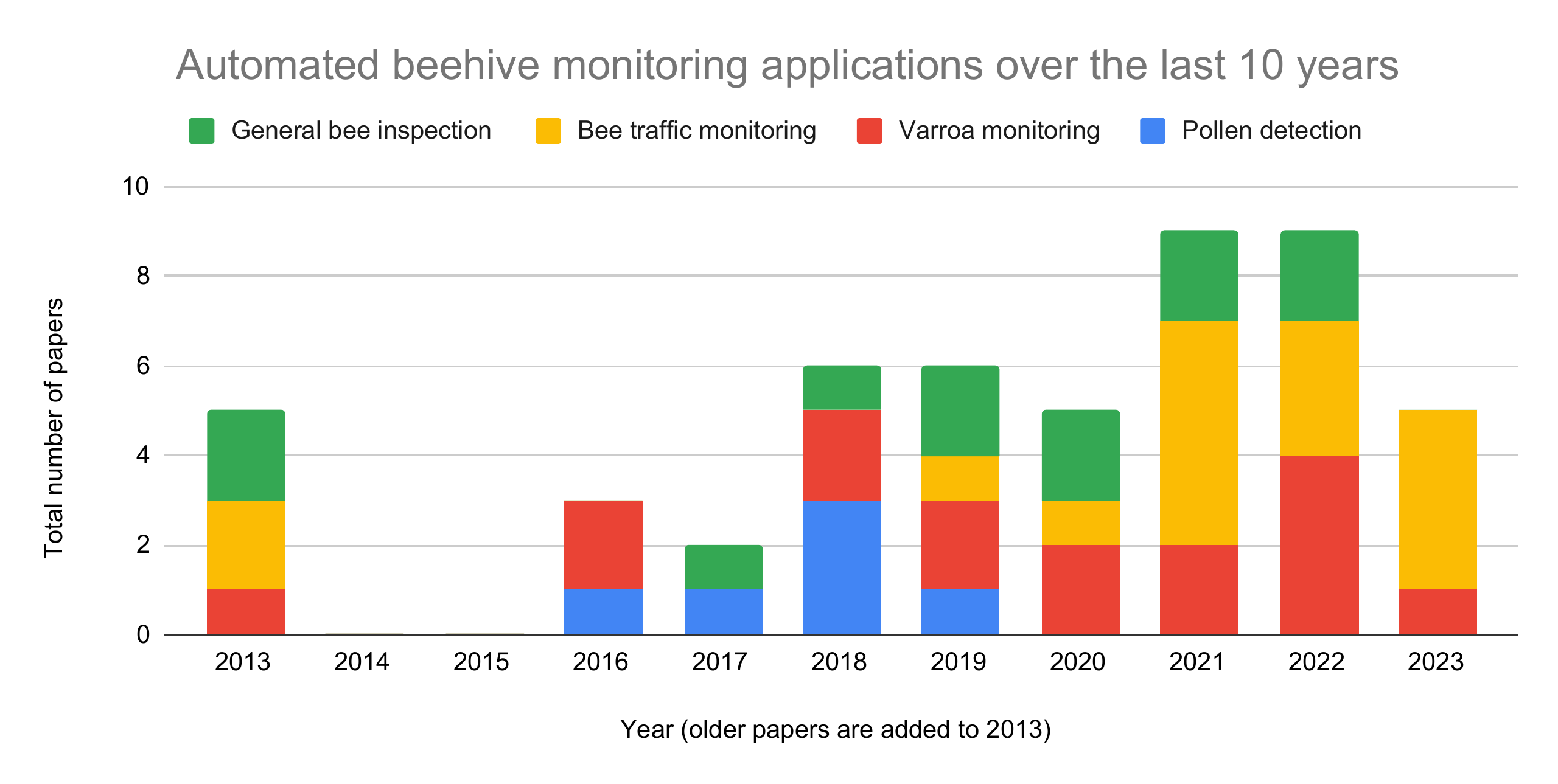}
 	\caption{Investigated automated beehive monitoring applications over the last 10 years}
 	\label{Fig:PaperCount}
\end{figure}

Considering the number of published papers and the application areas, the trends are shown in Fig. \ref{Fig:PaperCount}. It is apparent that from the 50 investigated papers, only five were published before the year 2013, included. From the year 2016, we could see a rise in the number of papers with a distinctive increase in the last five years, when in each of those years at least the same number of papers as before 2013, including, was published.

If we consider the application areas, we could see an increase especially in the bee traffic monitoring methods followed by the Varroa mite detection together with a small peak around 2018 with a dominance of the pollen detection methods. Considering the general bee inspection applications, which could not be categorized to the remaining categories, their number remains similar during the investigated period.

\begin{figure}[htb]
    \centering
 	\includegraphics[width=0.8\textwidth]{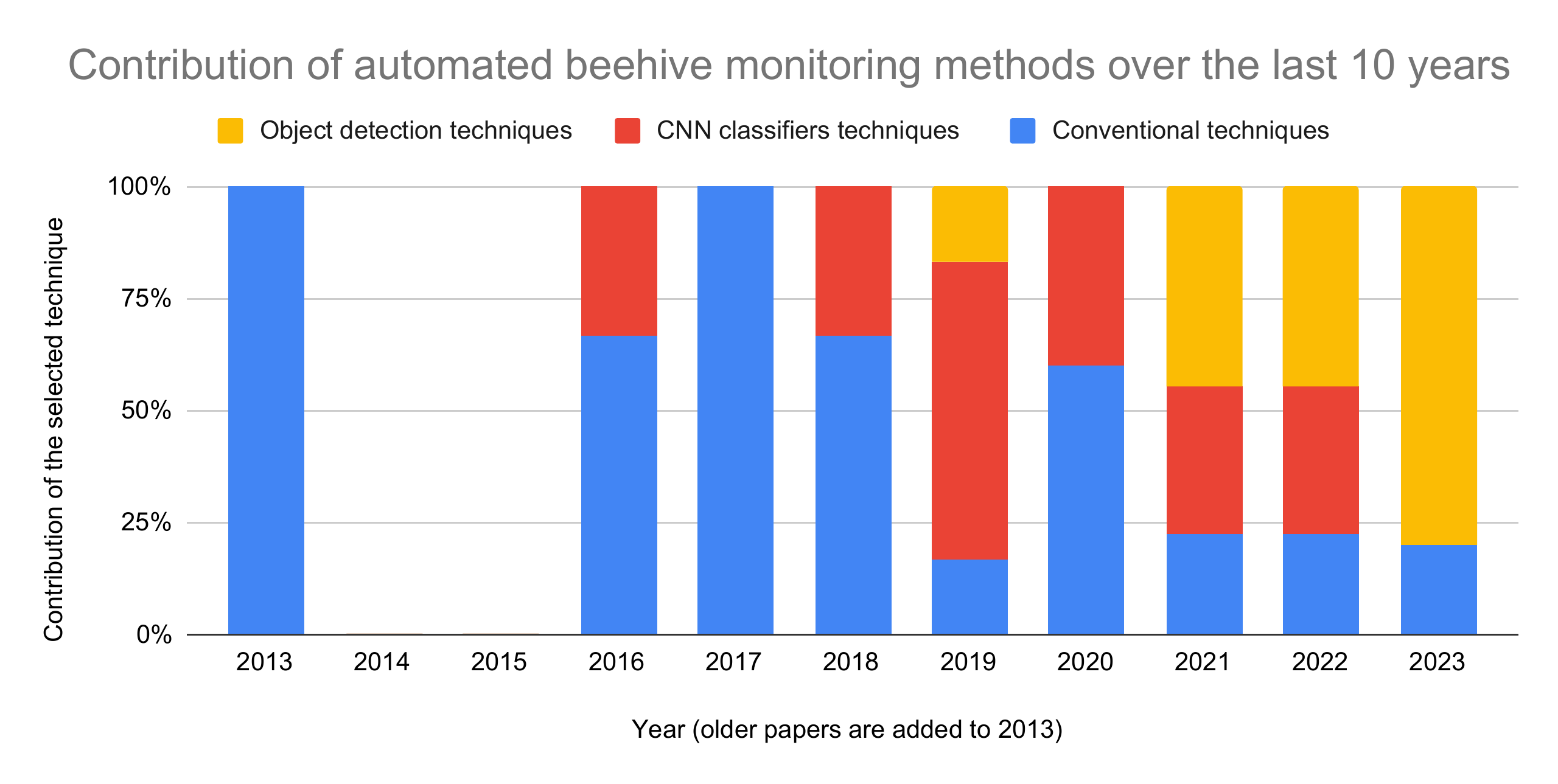}
 	\caption{Contribution of investigated automated beehive monitoring methods over the last 10 years}
 	\label{Fig:PaperContr}
\end{figure}

The composition of the investigated techniques over the last ten years is shown in Fig. \ref{Fig:PaperContr}. We could see, that the CNN based classifiers were used first in 2016 (one of the first succesful CNN based classifiers AlexNet was introduced in 2014), but the conventional techniques were dominant up until 2018 (considering only two investigated papers in the year 2017). The first use of object detection techniques investigated in our paper could be seen in the year 2019 with their dominance since the year 2021 (considering the introduction of first YOLO detector in 2015 and SSD detector in 2016).

\begin{figure}[htb]
    \centering
 	\includegraphics[width=0.8\textwidth]{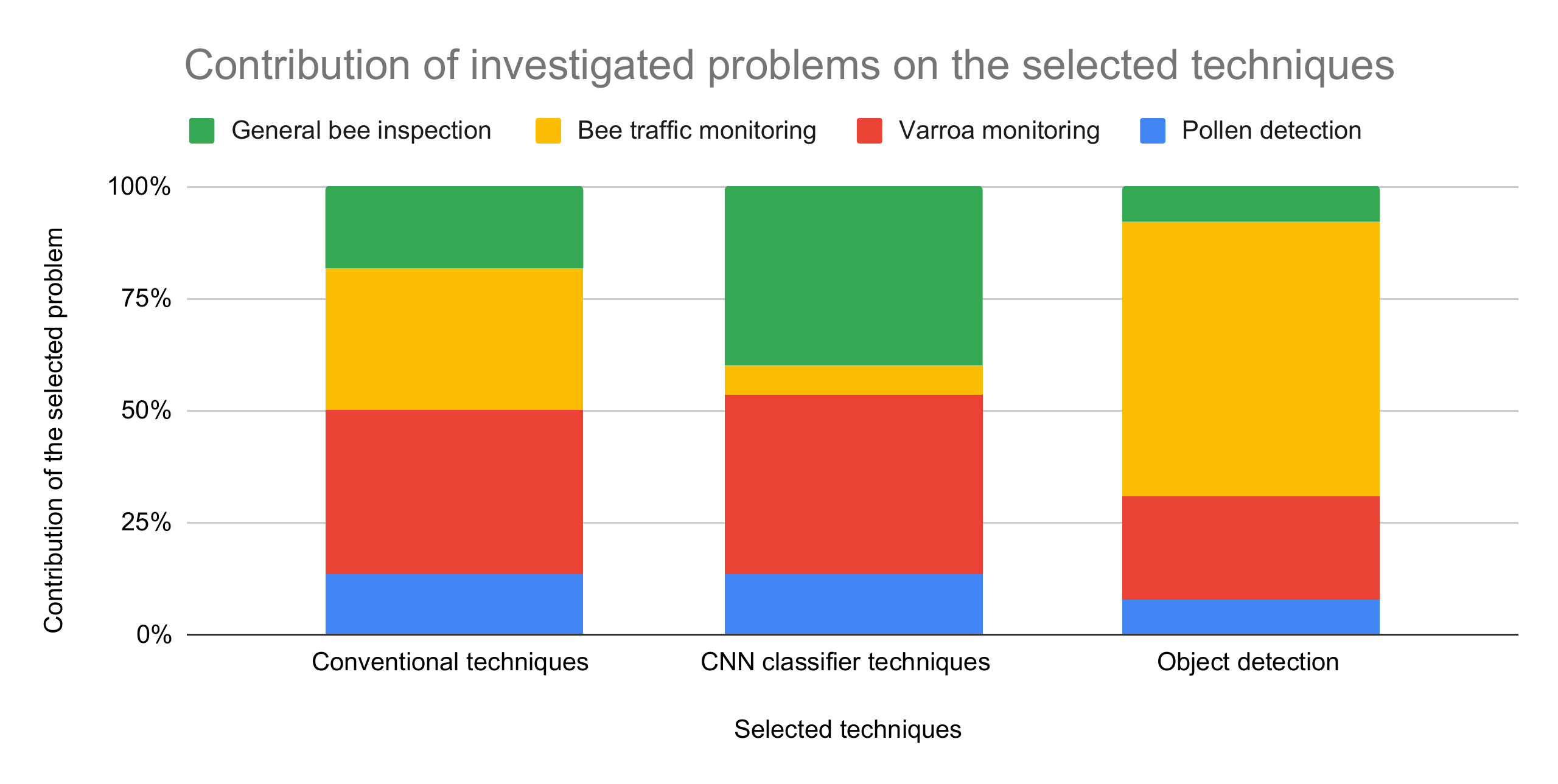}
    \includegraphics[width=0.8\textwidth]{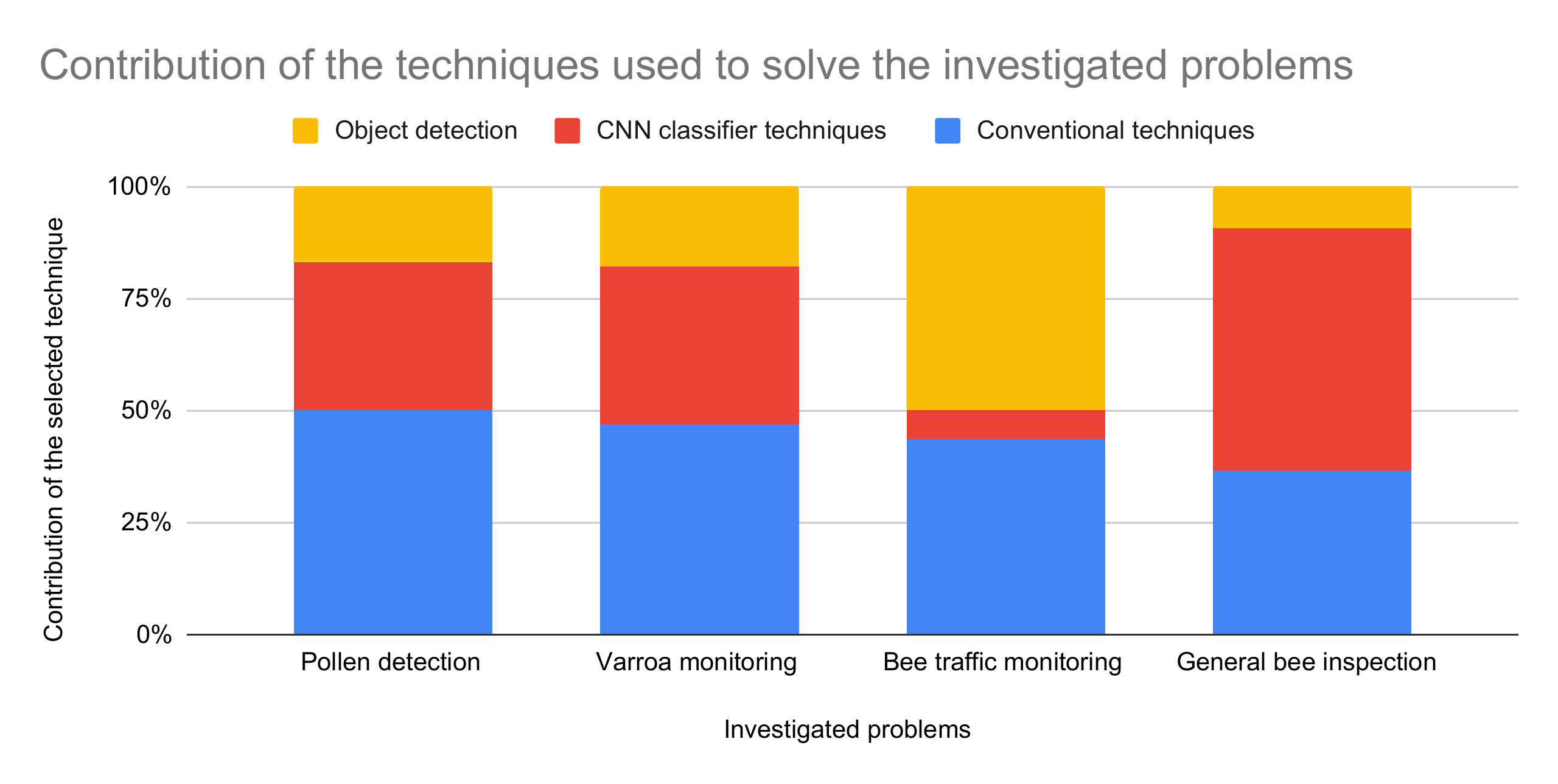}
 	\caption{Contribution of investigated problems on the selected techniques and vice versa over the last 10 years}
 	\label{Fig:PTTP}
\end{figure}

Contribution of the investigated problems on the selected techniques and vice versa is shown in Fig. \ref{Fig:PTTP}. Exploring the contribution of the investigated problems on the selected techniques, we could see that the conventional techniques are used almost in an equal ratio for all investigated problems. The CNN classifiers techniques are mostly used for the bee classification in general together with the Varroa monitoring. And finally that the majority of the object detection papers focus on bee traffic monitoring.

Regarding the contribution of techniques used to solve the investigated problems, it is apparent that a half of the pollen detection techniques are covered by conventional techniques followed by CNN classifiers and with a smaller contribution of object detectors. We could observe a similar situation in the Varroa monitoring. Another situation is the bee traffic monitoring problem, where a half of the applications is solved by object detectors followed by the conventional techniques and with a small contribution of the CNN classifiers. In the general bee inspection, the majority of problems is solved by the CNN classifiers and conventional techniques.

%\subsection{Possible future research directions}

Taking into account the investigated fields of application, the amount of papers and the trends in automated beehive monitoring research, we can see a rising number of research papers focusing on those applications. Those applications cover a wide variety of fields from pollen detection, Varroa mite detection, bee traffic monitoring, to brood cell inspection, bee and other insects species recognition, bee behaviour investigation and others.

Although the modern deep learning methods allow for an efficient solution of complex tasks such as image and speech recognition, machine translations and others, it is possible that further increasing of their accuracy is approaching computational and economical limits of the state-of-the-art algorithms and computational hardware as described in \cite{9563954}. With this in mind, lower computational power of embedded hardware and their usually better intelligibility, classical CV techniques might still be a good alternative in many applications. Nevertheless, we believe that many older studies using the conventional approach might be significantly improved by using deep learning, or object detection approaches, which also allow us to collect significantly more information from the given data using a single model.

In any case, there remains a significant obstacle to the deep learning approach – the lack of a well-made and annotated datasets in a high resolution. One solution could be joining some pre-existing datasets which could be difficult due to the different environments, or to create a new complex dataset made by long-term in-field measurement containing sufficient amounts of bees in various states (pollen-bearing, Varroa mite infestation…), drones, or even the bee queens. It could be interesting to also cover other sensory data such as sound, in-hive temperature and humidity, beehive weight or others. This would allow development of complex deep learning-based systems, which could be used for long-term bee colony monitoring and to recognize potentially dangerous situations.

A great opportunity for those systems is the broad range of available embedded devices, such as the Raspberry Pi family or the NVIDIA Jetson range. These devices are widely used, relatively cheap and with plenty of accessories, such as cameras or sensors. In the Raspberry Pi case, computational power is too weak  for efficient training of neural networks, but sufficient for real-time image processing using selected pre-trained models. Together with the well accessible 3D print technologies, this allows for rapid prototyping and brings good customization possibilities for various applications. Finally, it also allows for more involvement of the beekeeping community members, who previously might not have had access to the widely available state-of-the-art beehive monitoring devices.

Exploring the described research trends, we could see a significant rise in the application of the object detector techniques together with a rise in the number of papers covering bee traffic monitoring. We have come up with two hypothesis. The first hypothesis is that the pollen and Varroa mite detection problems are sufficiently solved mostly by the state-of-the-art CNN classifiers (but partially also with the conventional techniques), so the research moves to the more complicated field of bee traffic monitoring and motion pattern recognition. The second hypothesis is that the bee traffic monitoring problem is often solved in a more complicated environment, like the entrance to the beehive seen from above, with a large number of moving bees. This would be complicated to solve due to the needed segmentation of the individual objects, which would have to be done using conventional techniques, or an additional CNN classifiers. On the other hand, performance of the state-of-the-art object detectors such as SSD or YOLO allows a real-time processing on the embedded devices.

Applications like Varroa mite monitoring and pollen detection could be simplified by setting a proper scene with a non disturbing background and potential physical segmentation of the individual bees on the captured scene. We would even assume, that under a presumption of an appropriate scene and simplifying the problem to multi-class classification, the CNN based classifiers might be more suitable for those tasks. If the current trend continues and the usage of conventional techniques stagnates, it is likely that the contribution of those techniques will slowly decline in the favor of CNN based classifiers and object detectors.

%\clearpage

\section{Conclusion}\label{Sect:Concl}

In this paper, we presented an overview of 50 investigated research papers related to automated beehive monitoring applications with a focus on machine vision. For the purpose of this paper, we divided the surveyed papers into three categories according to the used technique (conventional methods, CNN based classifiers and object detectors) and further into four categories according the field of application (pollen detection, Varroa mite detection, bee traffic monitoring and bee inspection in general). Sections focusing on the selected technique are prefaced with a brief theoretical introduction and motivation.

We also discuss the problem of available bee datasets, which are crucial for CNN based systems and we emphasize a need for more complex high-quality datasets for developing other applications in this field.  Most of the presented techniques could be implemented in an embedded device, which could be used for in-field beehive monitoring, or research purposes. We also describe and discuss the research trends in this field based on the surveyed papers and the suitability of the described techniques for the investigated fields of application.

\section*{Acknowledgement}

The work was further supported by the grant number FEKT-S-23-8451 "Research on advanced methods and technologies in cybernetics, robotics, artificial intelligence, automation and measurement" from the Internal science fund of Brno University of Technology.

%\appendix
%\section{Overview table of the investigated papers}

%Appendix sections are coded under \verb+\appendix+.

%\verb+\printcredits+ command is used after appendix sections to list 
%author credit taxonomy contribution roles tagged using \verb+\credit+ 
%in frontmatter.

\printcredits

%% Loading bibliography style file
% \bibliographystyle{model1-num-names}
\bibliographystyle{cas-model2-names}

\end{document}